\documentclass[journal]{IEEEtran}
\usepackage{graphicx} 
\usepackage{enumitem}
\usepackage{hyperref}

\title{A Novel Collaborative Framework for Efficient Synchronization in Split Federated Learning over Wireless Networks}

\begin{document}

\author{Haoran~Gao,
Samuel~D.~Okegbile,~\IEEEmembership{Member,~IEEE,}
and~Jun~Cai,~\IEEEmembership{Senior~Member,~IEEE,}
\thanks{H. Gao and J. Cai are with the Department of Electrical and Computer Engineering, Concordia University, Montreal QC H3G 1M8, Canada (Emails: jun.cai@concordia.ca; haoran.gao@mail.concordia.ca).}
\thanks{S. D. Okegbile is with the Department of Physics and Computer Science, Faculty of Science, Wilfrid Laurier University, Milton ON L9T 5E1, Canada (Email: sokegbile@wlu.ca).}}

\maketitle

\begin{abstract}
Split Federated Learning (SFL) offers a promising approach for distributed model training in wireless networks, combining the layer-partitioning advantages of split learning with the federated aggregation that ensures global convergence. However, in heterogeneous wireless environments, disparities in device capabilities and channel conditions make strict round-based synchronization heavily straggler-dominated, thereby limiting both efficiency and scalability. To address this challenge, we propose a new framework, called Collaborative Split Federated Learning (CSFL), that redefines workload redistribution through device-to-device collaboration. Building on the flexibility of model partitioning, CSFL enables efficient devices, after completing their own forward propagation, to seamlessly take over the unfinished layers of bottleneck devices. This collaborative process, supported by D2D communications, allows bottleneck devices to offload computation earlier while maintaining synchronized progression across the network. Beyond the system design, we highlight key technical enablers such as privacy protection, multi-perspective matching, and incentive mechanisms, and discuss practical challenges including matching balance, privacy risks, and incentive sustainability. A case study demonstrates that CSFL significantly reduces training latency without compromising convergence speed or accuracy, underscoring collaboration as a key enabler for synchronization-efficient learning in next-generation wireless networks.
\end{abstract}


\section{Introduction}
Next-generation wireless networks increasingly rely on edge intelligence, which imposes stringent requirements on distributed machine learning frameworks. These frameworks are required to function under low latency and high reliability constraints that define next-generation communications. Federated Learning (FL) and Split Learning (SL) represent the main forms of distributed learning in wireless networks, supporting training without centralizing raw data. However, FL places substantial computational demands on end devices, whereas SL alleviates device workload through task offloading but suffers from high communication overhead and slow convergence~\cite{d2}. These limitations constrain their applicability in resource-constrained wireless networks.

Split Federated Learning (SFL) stands out as well suited to wireless networks, combining the aggregation capability of FL with the layer-partitioning of SL to achieve a balanced trade-off between device computation and wireless communication~\cite{t2}. In conventional SFL setups, the server must wait until all participating devices complete their training and smash data transmissions before proceeding to the next stage of training. This strict round-based synchronization preserves model consistency but renders the training straggler-dominated, as heterogeneous computing capacities across devices determine the overall system efficiency. In wireless environments, channel heterogeneity further aggravates this bottleneck. Differences in link quality and transmission rate cause some devices to take much longer to upload, thus the training is delayed by the slowest link, leaving faster devices idle and slowing convergence~\cite{g1}. The impact becomes even more pronounced in large-scale deployments, where the growing number of participants compounds the synchronization delays and further constrains system efficiency.

Motivated by these challenges, this paper presents a new framework, called Collaborative Split Federated Learning (CSFL), that mitigates the impact of bottleneck devices through device-to-device collaboration. In contrast to conventional approaches that adjust participation sets~\cite{n9}, relax synchronization schedules~\cite{s1, j1}, or reallocate computation placement~\cite{9}, CSFL employs peer-to-peer collaboration without external assistance. Leveraging the flexibility of SL, the proposed framework allows efficient devices, once completing their own forward propagation, to serve as relays that undertake the computations beyond the capabilities of bottleneck devices. In this way, bottleneck devices process only part of the model before handing over to their efficient counterparts, enabling collaboration, synchronized data transmission, and better utilization of available resources. The main contributions of this paper are summarized as follows.
\begin{itemize}
    \item  We are the first to exploit device-to-device collaboration to overcome synchronization issues, introducing the CSFL for synchronization in wireless networks.
    \item We provide an in-depth analysis of the requirements and challenges of implementing CSFL in wireless networks, detailing its architecture and workflow while highlighting key issues from multiple perspectives.
    \item We validate the effectiveness of CSFL through representative evaluations against common distributed learning frameworks, and we conclude with open research directions that may guide future developments in this area.
\end{itemize}

The remainder of this paper is organized as follows. Section II reviews the related work. Section III introduces the system model together with its major components, the associated design challenges, and the key techniques that enable the proposed framework. Section IV reports the simulation results. Finally, Section V concludes this research.

\section{Related Works}
Synchronization has long been a critical limitation in multi-device distributed learning, as the overall process is often delayed by the slowest participants. Early investigations started from FL, where global aggregation required all devices to complete local training before proceeding. To mitigate this issue, researchers proposed methods such as device selection~\cite{n9} and semi-synchronous FL (SSFL)~\cite{s1}. These methods alleviated the impact of stragglers to some extent. However, device grouping commonly adopted static strategies and yielded inconsistent updates across groups. Semi-synchronous aggregation, situated conceptually between synchronous and asynchronous operations, also remained vulnerable to delays caused by slower devices at synchronization points. Thus, researchers further explored the use of external resources~\cite{9}, in which auxiliary servers or relay nodes parallelized transmission and aggregation. While this reduced the direct communication burden on the central server, the dependence on additional nodes inevitably increased system complexity and raised scalability concerns.

Unlike FL, SL does not involve cross-device aggregation, but synchronization remains, as the server must still wait for the forward propagation from devices to proceed~\cite{g1}. To enhance efficiency under this setting, asynchronous update mechanisms were proposed, allowing devices to continue training without strict coordination~\cite{j1}. However, this design often resulted in model staleness and inconsistency, since faster devices could dominate the training process.


Built on the foundation of SL, SFL incorporates an additional aggregation server for device-side aggregation while also retaining server-side aggregation, thereby combining the structural advantages of SL with the consistency benefits of FL. In its default form, SFL relies on centralized synchronization, where the server must wait until all participating devices complete their computations before proceeding to the next stage, as illustrated in Fig. 1 through the forward and backward propagation between devices, the server, and the aggregation server~\cite{t2, g1}. Another synchronization approach is the peer-to-peer mode, where the server assigns each client a link to its predecessor, enabling devices to synchronize directly by downloading model weights. While this reduces the server’s involvement in intermediate steps, it requires the model updates to circulate through the entire chain of devices. As a result, the process is highly vulnerable to unstable network conditions and accumulative delays, and its scalability is limited since any weak link can disrupt the overall synchronization~\cite{d2}.

To further address synchronization limitations in SFL, \cite{t2} also proposed a sequential update order inspired by asynchronous training. In this protocol, the server accepts each device’s contribution in arrival order and updates the model immediately. While this reduces synchronization delay, the lack of aggregation weakens the consistency benefits of FL and gives early devices disproportionate influence, which may bias the model.

In summary, although existing studies could alleviate the impact of stragglers to some extent, but they still suffer from limited consistency and increased system complexity. In contrast, our proposed CSFL framework redefines workload allocation through device-to-device collaboration, enabling efficient devices to weaken the dominance of stragglers over training latency while preserving synchronous consistency and resource efficiency, thereby making it better suited to the demands of wireless networks.

\begin{figure}[!t]
  \begin{center}
  \includegraphics[width=3.4in]{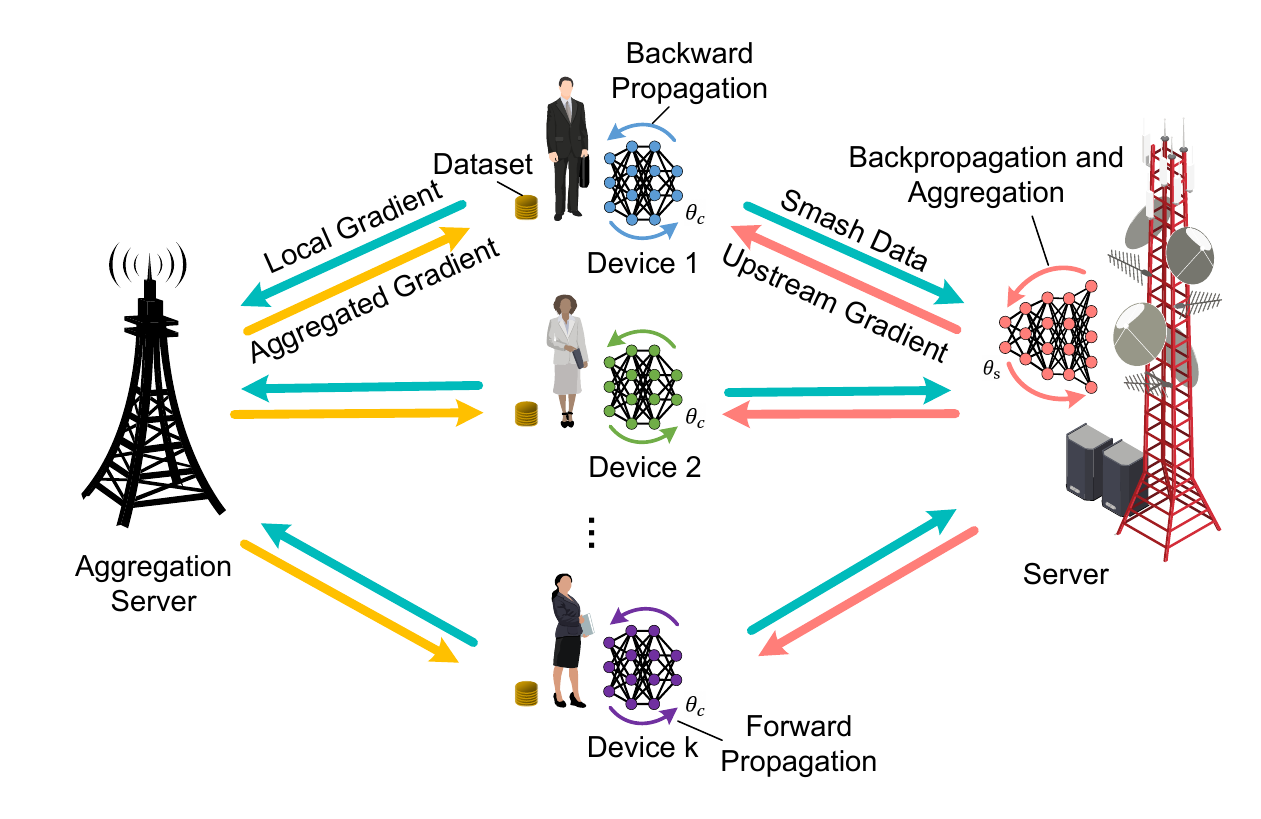}
  \caption{Standard Split Federated Learning Framework.}
  \end{center}
\end{figure}

\section{An Overview of CSFL Framework}
In this section, we propose a new CSFL framework to capture the synchronization requirement of SFL.

\begin{figure*}[!t]
  \begin{center}
  \includegraphics[width=5.78in]{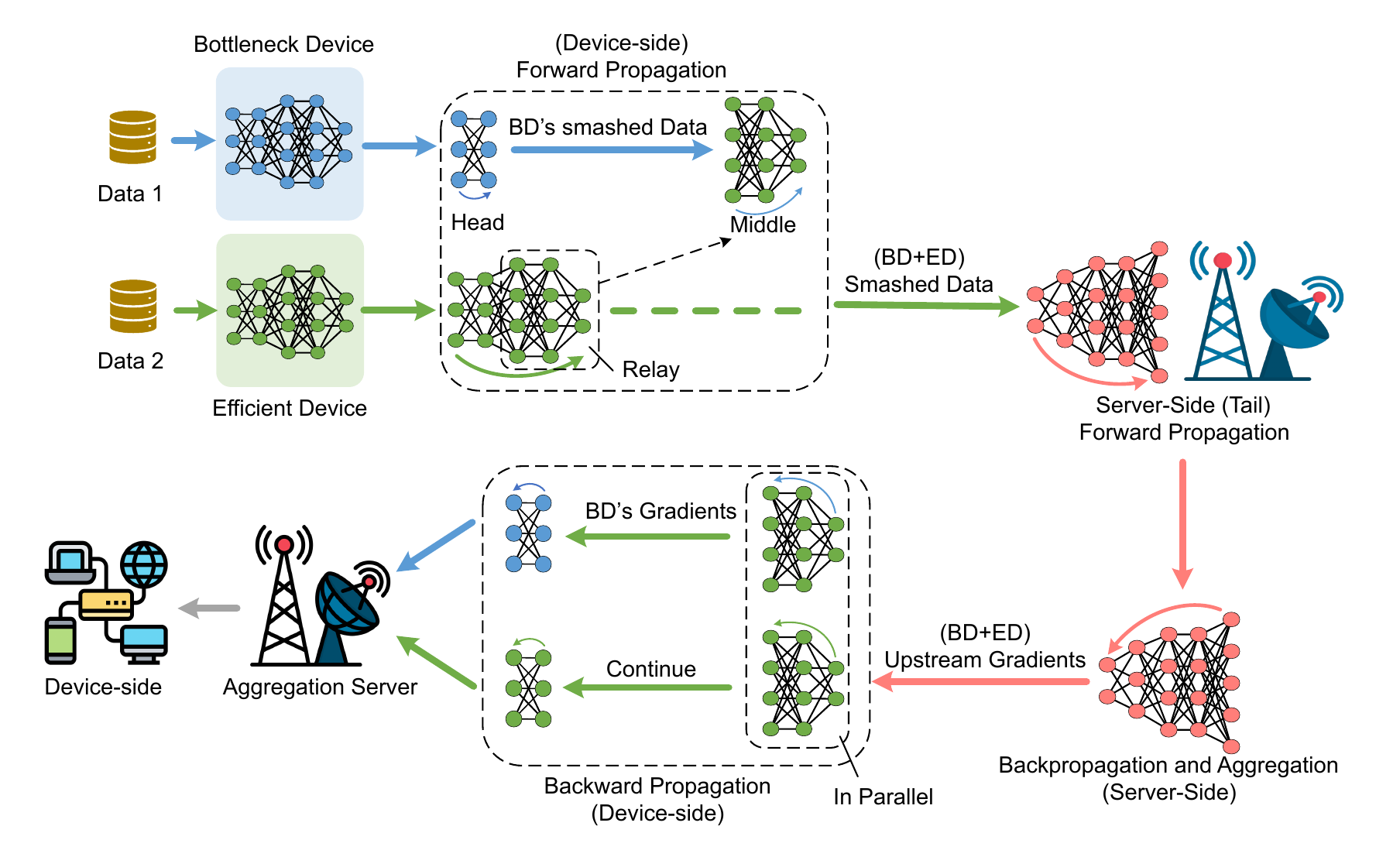}
  \caption{CSFL enables efficient devices to assist bottleneck devices, thereby enhancing the overall system's computational efficiency. The efficient devices will select the unfinished model layers to assist based on the portions of the model that the bottleneck devices have already completed.}
  \end{center}
\end{figure*}

\subsection{System Architecture}
The proposed CSFL framework is designed to ensure both model consistency and fairness while enabling an efficient and resource-optimized training environment. Its core mechanism is to classify devices as either efficient or bottleneck
based on performance profiling of their configurations, and then balance the load between efficient and bottleneck devices to reduce synchronization waiting time and improve resource utilization. 

Note that in SFL, a fundamental requirement is that all devices adopt a common cut layer~\cite{d2, t2}. Heterogeneous cut layers would fragment the feature space, lead to inconsistent smashed data dimensions across devices, cause misaligned gradient flows, increase system complexity and make aggregation infeasible. Thus, CSFL naturally inherits this requirement, as illustrated in Fig. 2 where efficient devices act as relays to assist bottleneck devices in the forward propagation. This structure, similar to a three-layer U-shaped SL architecture~\cite{l4}, allows the bottleneck devices handling the front-end portion of the forward propagation, the efficient devices processing the middle portion, and the server executing the tail-end portion.

In the forward propagation process, the system is divided into two stages, with adjustments made according to the different configurations of devices. In the first stage, all devices perform forward propagation using their own data. However, due to the more abundant resources and better configurations of the efficient devices, such devices complete the client-side part of the forward propagation before the bottleneck devices and proceed to the second stage.

In the second stage, once the efficient devices complete the forward propagation, the bottleneck devices pause their computations and pass the completed model layer information to the efficient devices. The efficient devices then determine the partition point within their own models, identifying the corresponding unfinished layers for further processing. Meanwhile, the bottleneck devices transmit the intermediate smashed data generated from the completed portion to the efficient devices via wireless channels (to distinguish it from the smashed data that will later be transmitted to the server, we refer to this as intermediate smashed data). The efficient devices will input this intermediate smashed data into the remaining selected model layers to complete the subsequent forward propagation. During this assistance period, the bottleneck devices may perform auxiliary tasks, such as enhancing the security or privacy protection mechanisms of the system and adjusting the batch-size for the next round of training to balance computational resources and improve model performance. After completing the relay computation, each efficient device uploads both its own smashed data at the cut layer and the relayed smashed data computed for its matched bottleneck device. The server then processes these inputs separately in the training pipeline.

Leveraging the superior computational resources of efficient devices together with the partitioning flexibility of SL accelerates the forward propagation of bottleneck devices, ensuring that their final smashed data is transmitted in synchrony with that of efficient devices. Once this synchronized forward process is completed, the server computes the gradients for its own layers and performs the parameter updates. The upstream gradients at the cut layer are then sent back to the efficient devices. Each efficient device uses these gradients to perform backpropagation through both its own layers and the layers it has relayed on behalf of its bottleneck device. The resulting gradients at the partition point are subsequently returned to the bottleneck devices, who continue the backpropagation through their local layers until completion. To ensure consistency, a separate aggregation server could be employed to aggregate the client-side model updates before synchronization with the main server. Since only smashed data are transmitted throughout the process, CSFL inherits the privacy-preserving property of SL, substantially mitigating concerns for ordinary devices~\cite{d2}. Stronger adversarial scenarios will be discussed in a later section.

CSFL retains the convergence characteristics and accuracy guarantees of conventional SFL. Its design leaves the optimization objective and effective batch size unchanged, merely redistributing computation and communication within each round. As a result, collaborative pairing primarily influences wall-clock training efficiency, while the convergence trajectory measured in training rounds remains unaffected. 

Generally, the key processes of the CSFL framework include layer splitting selection, efficient-bottleneck matching, determination of partition points and aggregation.

\begin{itemize}
    \item \textbf{Layer Splitting Selection}: The selection of split layers determines the division of computation between the client and the server, specifying which layers are executed locally and which are offloaded to the server. In standard SL, this choice shapes the balance between local computation and communication, since deeper cuts reduce transmission load at the cost of greater computational load on the client, while shallower cuts have the opposite effect. In CSFL, split layer selection plays an important role, as it not only affects this balance between computation and communication but is also closely related to the ability of efficient devices to act as relays for bottleneck devices. In particular, the chosen cut layer defines the amount of data transmitted and the additional computational load on efficient devices during assistance, which consequently affects overall resource utilization and training latency. The design objective is to strike a balance that maximizes system efficiency while ensuring all devices remain synchronized.
    \item \textbf{Efficient-Bottleneck Matching}: In CSFL, system efficiency is closely related to how efficient devices are paired with bottleneck devices. Proper matching enables the computational advantages of efficient devices to be utilized, thereby reducing synchronization delays and improving resource utilization. Conversely, mismatched pairing may result in underutilization or overload of efficient devices, while bottleneck devices may still limit the training process. The pairing process follows a criterion, whereby devices are matched if the anticipated cooperative latency falls below the latency of independent execution, with both computational disparity and communication overhead considered. In practice, the numbers of efficient and bottleneck devices may not always be equal, which can lead to imbalance. To remain effective under such conditions, the matching strategy should be flexible enough to ensure fairness among devices while still reducing overall synchronization latency.
    \item \textbf{Determination of Partition Points}: The selection of partition points is based on the device matching results. Unlike the previously discussed split layer selection, partitioning involves choosing the relay model between devices. Specifically, the relay model for efficient devices usually excludes part of the client model, as bottleneck devices independently handle the initial computations. The partition point therefore indicates the number of layers completed by the bottleneck device and the unfinished layers that efficient devices will process, while guiding efficient devices to preserve the necessary layers for assistance and avoid redundant operations. The choice of partition point directly affects the computational load on efficient devices and the amount of data transmitted, and thus plays an important role in maintaining resource efficiency and avoiding unnecessary training delays.
    \item \textbf{Aggregation}: The aggregation process in the CSFL framework follows the same principle as in SFL, integrating client model weights to maintain model consistency and synchronization. In CSFL, it plays an essential role in enabling efficient devices to assist bottleneck devices, since consistent parameter updates are necessary for their collaboration. This ensures that the contributions of all devices remain aligned with the global training objective, supporting both accuracy and stable system performance.
\end{itemize}

\subsection{Design requirements and challenges}
Clearly, the previous analysis indicates that implementing the CSFL framework is not straightforward. Next, we will provide a detailed discussion of the key design requirements and challenges involved in realizing a CSFL-enabled framework.

\begin{itemize}
    \item[1)] \textbf{Device Retrieval Complexity}: Device retrieval refers to the process by which efficient devices identify suitable bottleneck devices for assistance. In the simplest case of two clients, the procedure is straightforward, as the efficient device only needs to select its counterpart for matching. In realistic scenarios, the number of clients connected to a server is typically much larger, which makes retrieval considerably more involved. Efficient devices must evaluate a wider set of candidates, and this increases both the communication interactions required to exchange status information and the computational effort needed to make pairing decisions. As the client population grows, the set of possible matching decisions expands rapidly, which complicates the optimization process and raises the requirements on the decision-making strategy. Thus, the challenge is to narrow this decision space in a systematic way and to improve the efficiency of device matching, so that the scalability of the CSFL framework can be maintained even under large and heterogeneous deployments.
    \item[2)] \textbf{Model Depth Validation}: The depth of the model is directly related to the determination of the split layer. In conventional SFL, this choice is typically made by balancing computational demand and communication cost~\cite{w3}. The split point determines how much of the workload is retained at the device and how much is offloaded to the server. However, in CSFL, the decision becomes more intricate because it involves multi-level load balancing. Efficient devices must allocate resources to both their own computations and the relay tasks they perform for bottleneck devices. A deeper split layer shifts more computation to efficient devices, while a shallower split lightens the load on efficient devices and leaves more work with bottleneck devices. Relay overhead also needs to be taken into account, as the size of the exchanged data directly affects latency and bandwidth utilization. These interacting factors make split layer selection a central design challenge, requiring careful coordination to maintain both efficiency and scalability.
    \item[3)] \textbf{Balancing in Matching}: Matching constitutes a key design element in the proposed framework and remains among the most challenging aspects. Since it is interdependent with other components of the system, any limitation in those stages may propagate to the matching process. Beyond these dependencies, matching also entails several inherent challenges:

    \begin{itemize}
        \item[\textit{i.}] \textit{Global synchronization}: Each group in the matching process consists of one efficient device and one bottleneck device. Local synchronization within a group is achieved through collaborative between the paired devices, but achieving global synchronization requires coordination across all groups to minimize overall latency. The challenge lies in balancing not only the synchronization efficiency of individual pairs but also the consistency of synchronization times across the entire system.
        \item[\textit{ii.}] \textit{Device Number Imbalance}: In real-world scenarios, there may be more efficient devices than bottleneck devices, or vice versa. This imbalance means that the matching process must account for both matched groups and the presence of unmatched devices. For example, when the number of bottleneck devices exceeds that of efficient devices (a more typical case), matching may need to decide whether to temporarily exclude some bottleneck devices from the current training round, or to allow certain efficient devices to assist multiple bottleneck devices in a one-to-many configuration. The number of bottleneck devices that each efficient device can support is not fixed and may vary depending on its computational and communication resources. For the exclusion case, if some bottleneck devices cannot be matched in one round, they should be rotated across rounds to ensure fairness and to prevent any device from being persistently excluded from the collaborative process, which results in a long-term optimization problem. Conversely, when efficient devices are more numerous, unmatched efficient devices can proceed with independent local training in that round.
        \item[\textit{iii.}] \textit{Computational resources}: Leveraging efficient devices to assist bottleneck devices introduces additional considerations for resource management. The residual resources of efficient devices, including input data size and computational capacity, may fluctuate as training progresses. Although efficient devices often complete their own tasks ahead of others, their remaining resources are not guaranteed to support extra simultaneous assistance for multiple devices. The frequency of assistance also needs to be regulated, since intensive support could lead to sustained high loads. In such cases, allocating efficient devices to independent training may provide a balanced alternative to ensure overall system stability.
        \item[\textit{iv.}] \textit{Wireless Communication}: In CSFL, device-to-device (D2D) communication is promising for the communications between efficient and bottleneck devices. In practice, however, D2D links is constrained by factors such as physical distance or interference. This creates challenges, where fluctuating channel conditions, signal interference, and bandwidth limitations influence both quality and reliability. Such variations in channel state information (CSI) and network environments need to be incorporated into the matching strategy, since the stability of D2D links directly affects the effectiveness of device collaboration.
        \item[\textit{v.}] \textit{Optimization}: Incorporating all these factors into the optimization process substantially increases its complexity. This multi-dimensional optimization requires balancing computing load, communication resources, and synchronization latency, while also accounting for uncertainties in a dynamic environment. Consequently, developing an optimization strategy that is both effective and practical remains an important challenge for CSFL system design.
    \end{itemize}
    \item[4)] \textbf{Privacy}: As discussed in the system model, smashed data replace raw inputs and provide a degree of privacy protection; however, in CSFL the assistance procedure requires sharing smashed data between devices, which can still contain information that adversaries may exploit. For example, model inversion attacks can leverage the residual information in smashed data to partially reconstruct the original input~\cite{z2}. In addition to this sharing risk, device-to-device relaying also introduces secondary exposure during transmission, as smashed data traverse D2D links before reaching the server, so each relay hop increases the opportunity for interception or misuse at the link or endpoint. The fundamental challenge is to balance the efficiency gains of such collaboration with the amplified risks of privacy leakage, ensuring that reductions in latency do not come at the cost of data confidentiality.
    \item[5)] \textbf{Incentive Mechanisms}: In practice, the absence of incentive mechanisms creates challenges for motivating efficient devices to assist bottleneck devices, particularly when collaboration involves unfamiliar participants. Efficient devices require sufficient motivation or rewards to offset the resource consumption and potential delays associated with assistance. Designing such mechanisms is non-trivial, as it involves fairly quantifying and rewarding contributions under heterogeneous conditions. The variability of device capabilities and the unpredictability of collaborative environments further complicate the definition of standardized metrics for compensation. Addressing these challenges is essential to ensure that efficient devices are consistently willing to participate in collaboration beyond long-term or pre-established partnerships.
\end{itemize}

\subsection{Key Techniques}
The following techniques are potential to address the specific requirements and challenges associated with the implementation of CSFL framework.

\begin{itemize}
    \item[1)]  \textit{Dimensionality Reduction Methodology}: The contraction of device retrieval dimensions implies a reduction in the action space, which involves the precise identification of bottleneck devices by efficient devices. The exclusion of efficient devices ensures that the action space consists solely of bottleneck devices, which not only reduces the action space but also establishes a foundation for more accurate device positioning and pairing in future iterations. A viable solution is to implement a cluster assignment strategy. Cluster assignment creates different clusters based on device resource profiles, ensuring that devices with similar configurations are grouped into the same cluster~\cite{okegbile2025fles}. Through this assignment strategy, efficient devices and bottleneck devices can be initially categorized, which not only reduces the search overhead for efficient devices but also improves training efficiency while capturing long-range dependencies.
    \item[2)]  \textit{Privacy Considerations}: Privacy preservation in CSFL involves a trade-off between efficiency and privacy. The primary concern arises from the fact that smashed data must be shared between devices during collaboration. This makes the protection of shared smashed data a fundamental requirement in CSFL. One practical approach is to confine such computations to hardware-protected environments, such as Trusted Execution Environments (TEEs)~\cite{99}, which establish secure regions where intermediate features can be processed without being disclosed to efficient devices or other untrusted entities. TEEs can also be combined with multi-party computation techniques to provide cryptographic protection for data exchange, thereby further strengthening confidentiality within the collaborative training process. Beyond this sharing risk, D2D transmission introduces a secondary exposure. To mitigate this, semantic communication~\cite{seman} can be employed to limit what can be inferred from transmitted features and thereby lower the likelihood of reconstruction attacks.
    \item[3)]  \textit{Dynamic Optimization Technique (DOT)}: DOT involves algorithms that can adjust system parameters in real-time within dynamic environments~\cite{w3}. Their core feature is the ability to optimize strategies based on the current state of the system and changes in external conditions, ensuring that the system maintains high efficiency under varying circumstances. In the CSFL framework, factors such as resource availability, network conditions, and device configurations frequently change. Deep Reinforcement Learning (DRL) can be integrated into these dynamic optimization techniques, enabling the system to learn optimal strategies through continuous feedback from the environment. DRL monitors these variations and provides real-time feedback to automatically adjust computational loads, communication overhead, and synchronization strategies, thereby optimizing model partitioning across different scenarios. 
    \item[4)]  \textit{Multi-Perspective Matching Structure}: The matching between efficient devices and bottleneck devices requires integrating multiple considerations rather than relying on a single metric. At the individual level, factors such as computing time and energy consumption determine each device’s capacity to participate in cooperation. At the pair level, the communication overhead and synchronization latency between the two devices define the efficiency of collaboration. At the global level, system-wide requirements such as overall latency and fairness constrain the feasible matching outcomes. Balancing these perspectives is non-trivial, since optimizing one dimension may come at the expense of another. Game-theoretic approaches may be particularly suitable in this context~\cite{qoe}, as they provide a systematic way to model interactions among devices and to establish stable pairing relationships under competitive conditions, thereby enhancing the robustness of the CSFL framework. In common modeling practices, one-to-one pairing is widely adopted, as it reflects the natural rationale that stronger devices assist weaker ones to enhance overall system efficiency, and thus provides a natural baseline for analysis. 
    \item[5)]  \textit{Contract-Based Incentive Mechanisms}: Beyond achieving stable matching, it is equally important to ensure that efficient devices remain motivated to assist bottleneck devices after pairing. Contract theory offers a systematic approach to designing incentive mechanisms that align individual benefits with system-wide objectives~\cite{incentive}. By establishing agreements that balance costs and rewards, cooperation can be sustained without undermining overall efficiency. Such mechanisms align individual incentives with system-wide performance goals, thereby promoting sustained collaboration within the CSFL framework.
\end{itemize}

\section{A Case Study of CSFL Framework}
To assess the practical value of CSFL under device heterogeneity, we present a focused case study on a distributed learning workload. The objective is to evaluate not only whether the proposed framework can sustain model accuracy, but also whether it can mitigate the synchronization bottlenecks that arise in heterogeneous environments. To this end, we demonstrate two complementary performances: classification accuracy (top-1 accuracy) versus training rounds, which characterizes the convergence behavior of the model, and end-to-end latency versus training rounds, which captures the overall efficiency of the training process from local computation through communication to server updates. These two performances jointly demonstrate both the learning effectiveness and the responsiveness of the system. All configurations are evaluated under identical conditions, using the same backbone and optimizer; the only distinction lies in the collaboration and matching strategy applied.

For comparison, we consider four baselines:
\begin{enumerate}[label=\arabic*)]
    \item \textbf{SL}: This framework represents a multi-device, single-server split learning framework where training proceeds without aggregation on either the device side or the server side. SL is served as a useful baseline to highlight the role and importance of aggregation in collaborative training.
    \item \textbf{SFL}: Building upon SL, it introduces aggregation at both the server and the client sides. Throughout our simulations, SFL is adopted as a key baseline to demonstrate that CSFL not only inherits the convergence guarantees of SFL but also achieves improved efficiency under heterogeneous conditions.
    \item \textbf{CSFL with random matching (CSFL-R)}: CSFL-R adopts the same structure as our proposed CSFL but forms one-to-one pairing in an arbitrary manner, without considering device heterogeneity or cooperative latency. While it retains the relay mechanism of CSFL, the lack of an optimized matching policy often leads to suboptimal collaboration. This variant is included as a baseline to demonstrate the necessity of adopting a well-designed pairing strategy.
    \item \textbf{CSFL with greedy matching (CSFL-G)}: CSFL-G represents a practical variant of our framework, in which the greedy strategy provides a simple yet effective matching mechanism without relying on complex global optimization. Each efficient device selects a bottleneck device by jointly considering computational load and communication delay, ensuring that cooperative pairs are formed whenever the expected collaborative latency is lower than independent execution. This makes CSFL-G serves as an important baseline for validating the CSFL concept.
\end{enumerate}

\begin{figure}[!t]
    \centering
    \includegraphics[width=0.48\textwidth]{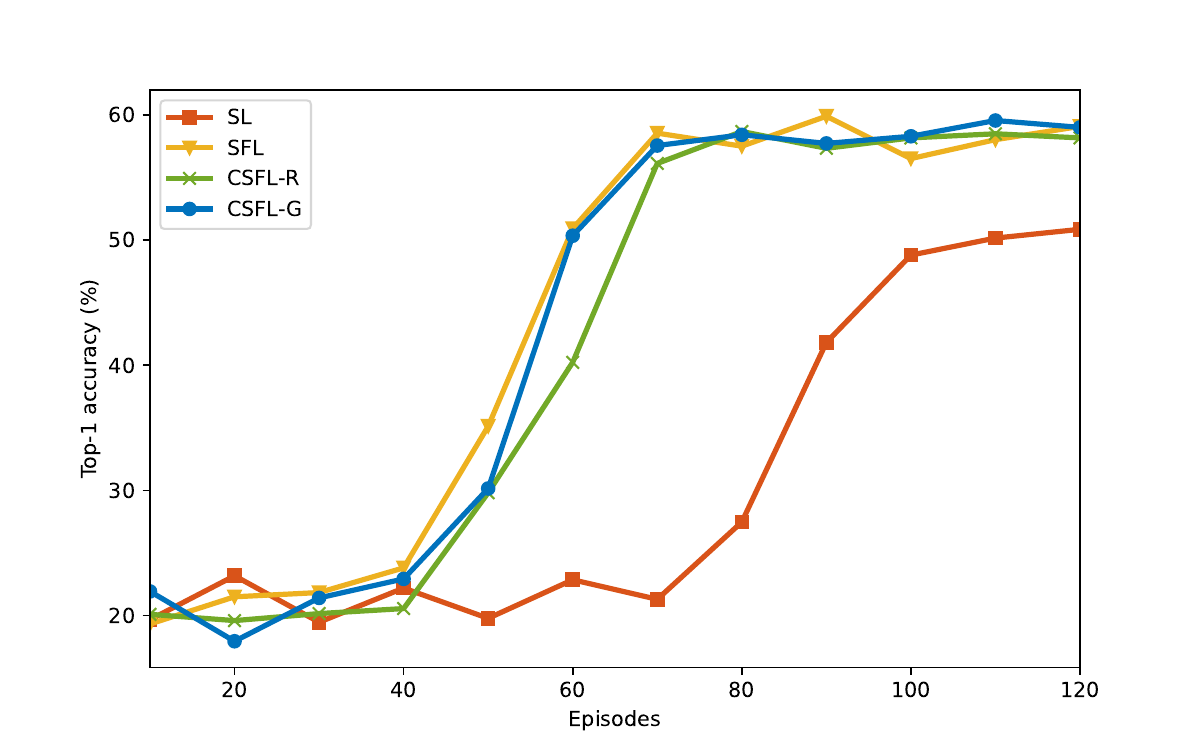} 
    \caption{Performance in terms of Top-1 accuracy.}
\end{figure}

\subsection{Simulation Settings}
To examine performance, we adopt the Amazon FullReview dataset\footnote{AmazonReviewFull dataset: https://www.kaggle.com/datasets/zhangjuefei\\/amazon-review-full}, a large-scale dateset for text classification that naturally introduces heterogeneity across devices. Its diversity in content and data volume makes it perfect for analyzing synchronization behavior and validating the efficiency of the CSFL framework.

In our implementation, the backbone model follows a TabTransformer-based design tailored for text classification on the Amazon FullReview dataset. On the device side, the model includes an embedding layer that maps categorical features into dense vectors, an input projection layer with ReLU activation, and the first three transformer encoder blocks. The smashed data generated at this stage are transmitted to the server, which hosts the remaining three transformer encoder blocks, a column-wise mean pooling layer, and a fully connected classification head with two hidden layers of size 512 and 256 (ReLU) followed by a softmax output layer with five classes. This partitioning scheme aligns with SL, where computationally intensive layers are executed on the server, while lightweight embedding and early encoding remain on the device side, thereby ensuring cross-device consistency and enabling flexible collaborative training under CSFL.

\begin{figure}[!t]
    \centering
    \includegraphics[width=0.48\textwidth]{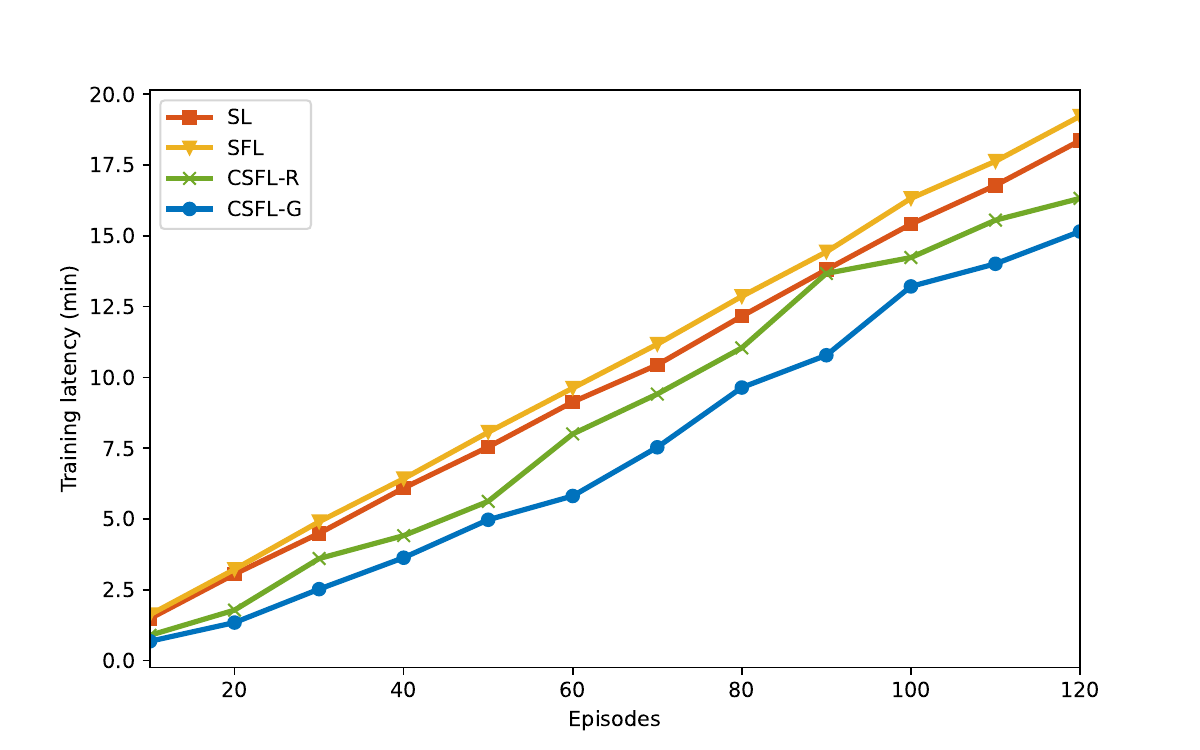} 
    \caption{Efficiency in terms of training latency.}
\end{figure}

\subsection{Evaluation Results}
We fixed the number of devices at 20 and allocated 500 samples to each device. For SFL, CSFL-R, and CSFL-G, aggregation is performed in every training round, which represents one complete forward–backward propagation cycle executed on a device. We first examined the top-1 accuracy to assess whether the collaborative mechanisms in CSFL affect convergence behavior and final performance. As shown in Fig. 3, both CSFL-R and CSFL-G achieve convergence rate and final accuracy comparable to those of standard SFL without collaboration, demonstrating that the incorporation of cooperation does not impair training effectiveness. In contrast, SL shows markedly inferior performance, as the absence of aggregation leads to significantly slower convergence. Moreover, without periodic aggregation, the model fails to benefit from knowledge integration across devices, essentially restricting training to the limited 500 samples available per device, which severely compromises generalization capability.

We further evaluated the system efficiency after a fixed 120 training rounds, using training latency as the primary metric, as illustrated in Fig. 4. Here, training latency is defined as the total training time, measured in minutes. The results show that CSFL-G, despite relying on a greedy rather than an optimal pairing strategy, already achieves a substantial reduction in latency. CSFL-R exhibits slight fluctuations due to the randomness in pairing, yet its overall latency remains lower than that of both SFL and SL. Among all methods, SFL incurs the highest latency, as the inclusion of aggregation steps introduces a small but non-negligible overhead compared with SL. Taken together, the analyses of both accuracy and training latency demonstrate the feasibility and efficiency of the proposed CSFL framework.
 
\section{Conclusions and Future Research Directions}
In this paper, we introduced a novel CSFL framework, showcasing its potential to enhance synchronization in SFL, thereby reducing overall training latency and facilitating more efficient resource utilization. We detailed the general architectural framework, highlighting essential design requirements and addressing key challenges. Additionally, we explored critical techniques crucial for implementing CSFL, supported by a comprehensive case study that demonstrates the framework's effectiveness in improving content generation. While our findings are promising, several important open issues remain that warrant further investigation.

\begin{itemize}
    \item[1)] \textbf{Effective cluster assignment algorithm}: The design of clustering and matching mechanisms that can adapt to dynamic device resources and network conditions remains a challenging problem. Most existing studies are conducted under static assumptions, whereas practical environments are inherently dynamic. Developing flexible and adaptive strategies will therefore be crucial to achieve stable and efficient collaboration.
    \item[2)] \textbf{Lightweight alternatives to TEEs}: While TEEs provide a secure means to protect feature-layer computations, their high energy consumption and hardware dependency limit large-scale adoption on mobile and edge devices. Exploring lightweight and portable alternatives, such as streamlined hardware modules or trusted virtualization environments, remains an important research direction.
    \item[3)] \textbf{Scalability and large-scale deployment}: Most evaluations of CSFL have been limited to small-scale scenarios. Demonstrating scalability to hundreds or thousands of devices, while maintaining convergence and efficiency, is critical for practical adoption. Validating CSFL in real-world applications, including large language models (LLMs), will be an important step forward.
\end{itemize}

\bibliographystyle{ieeetr}
\bibliography{reference}

\end{document}